\documentclass{article}

\usepackage[preprint]{neurips_2023}

\usepackage[utf8]{inputenc} % allow utf-8 input
\usepackage[T1]{fontenc}    % use 8-bit T1 fonts
\usepackage{hyperref}       % hyperlinks
\usepackage{url}            % simple URL typesetting
\usepackage{booktabs}       % professional-quality tables
\usepackage{amsfonts}       % blackboard math symbols
\usepackage{nicefrac}       % compact symbols for 1/2, etc.
\usepackage{microtype}      % microtypography
\usepackage{xcolor}         % colors
\usepackage{amsmath}
\usepackage{algorithm}
\usepackage{algpseudocode}
\usepackage{graphicx,subfigure} 
\DeclareMathOperator*{\argmin}{arg\,min}

\title{PS-FedGAN: An Efficient Federated Learning Framework 
Based on Partially Shared Generative Adversarial Networks For Data Privacy}

\author{%
  Achintha Wijesinghe\\
  Department of Electrical and Computer Engineering\\
  University of California, Davis\\
  Davis, CA \\
  \texttt{achwijesinghe@ucdavis.edu} \\
  \And
  Songyang Zhang \\
  Department of Electrical and Computer Engineering\\
  University of California, Davis\\
  Davis, CA \\
  \texttt{sydzhang@ucdavis.edu } \\
  \AND
  Zhi Ding \\
  Department of Electrical and Computer Engineering\\
  University of California, Davis\\
  Davis, CA \\
  \texttt{zding@ucdavis.edu} \\
}

\begin{document}

\maketitle

\begin{abstract}
Federated Learning (FL) has emerged as an effective learning paradigm for distributed computation owing to its strong potential 
in capturing underlying data statistics while preserving data privacy. However, in cases of practical data heterogeneity among 
FL clients, existing FL frameworks still exhibit deficiency in capturing the overall feature properties of local client data 
that exhibit disparate distributions. In response, generative adversarial networks (GANs) have recently been exploited in FL to address data heterogeneity since GANs can be integrated for data
regeneration without exposing original raw data. Despite some successes, existing GAN-related FL frameworks often incur heavy communication cost and also elicit other privacy concerns, which limit their applications in real scenarios. To this end, this
work proposes a novel FL framework that requires only
partial GAN model sharing. Named as \textbf{PS-FedGAN}, 
this new framework enhances the GAN releasing and training mechanism to address heterogeneous data distributions
across clients and to strengthen privacy preservation at
reduced communication cost, especially over wireless networks. 
Our analysis demonstrates the convergence and privacy benefits
of the proposed PS-FEdGAN framework. Through experimental results based on several well-known benchmark datasets, our proposed PS-FedGAN shows great promise to tackle FL under non-IID client data distributions, while securing data privacy and lowering communication overhead. 
\end{abstract}

\section{Introduction}

% In the past decade, Machine Learning (ML) technologies, especially Deep Learning (DL), have gained enormous interest in both academics and industries. Moreover, Due to the data-driven nature and the ability to tackle with non-linearity of the problem, ML/DL-based methods have shown their superiorities compared to traditional approaches in many areas such as Nature Language Processing (NLP), Computer Vision (CV), and even wireless communication. Despite the superior performances of ML, one of the major drawbacks is the need for data centralization which may not be good news for distributed systems.  Such systems require distributed users to share data with a centralized server raising many privacy as well as communication concerns.

Federated Learning (FL) presents an exciting framework for distributed and collaborative learning while preserving users' data privacy~\cite{Fed1, FL}. As an emerging field of artificial intelligence, the 
full potential of FL requires us to effectively address a myriad of challenges, 
including heterogeneity of data distribution, data privacy consideration, and communication efficiency. 
Among many existing FL schemes to handle data heterogeneity, generative adversarial networks (GANs)~\cite{GAN1} based approaches have recently drawn substantial interest owing to their ability to regenerate data statistics without sharing raw data. Despite much success, GAN-based FL is also prone to several shortcomings, such as poor privacy protection and heavy communication redundancies, caused by full model or synthetic data sharing~\cite{IJCAL,FL_syn}.

Many studies such as ~\cite{attack1, Attack2} have shown that  traditional FL methods may be vulnerable and can  cause sensitive information leakage. Similar concerns happen in GAN-based FL.
As discussed in~\cite{gan_leaks}, releasing GAN or synthetic data may lead to critical privacy issues. 
Specifically, GANs are prone to reconstruction attacks and membership inference attacks, where an attacker aims to regenerate data samples and check the usage of a certain data sample ~\cite{logan}, respectively. 
To address related privacy concerns, differential privacy (DP)~\cite{DP1} has been studied in GAN-based FL \cite{DP-GAN, IJCAL}. However, the major drawback of DP is the tradeoff between privacy and utility where a privacy budget is introduced to balance performance and privacy. Thus,
to leverage the performance of downstream tasks, an infinite privacy budget is often picked, 
thereby providing no privacy. On the other hand,  GAN-based FL also suffers from communication inefficiency 
due to the large size of shared models and limited communication resources in many realistic networking
scenarios. We provide a more detailed literature review in the Appendix in view of the page limit. 

\begin{figure}[htb]
  \centering
  \includegraphics[width=0.78\columnwidth]{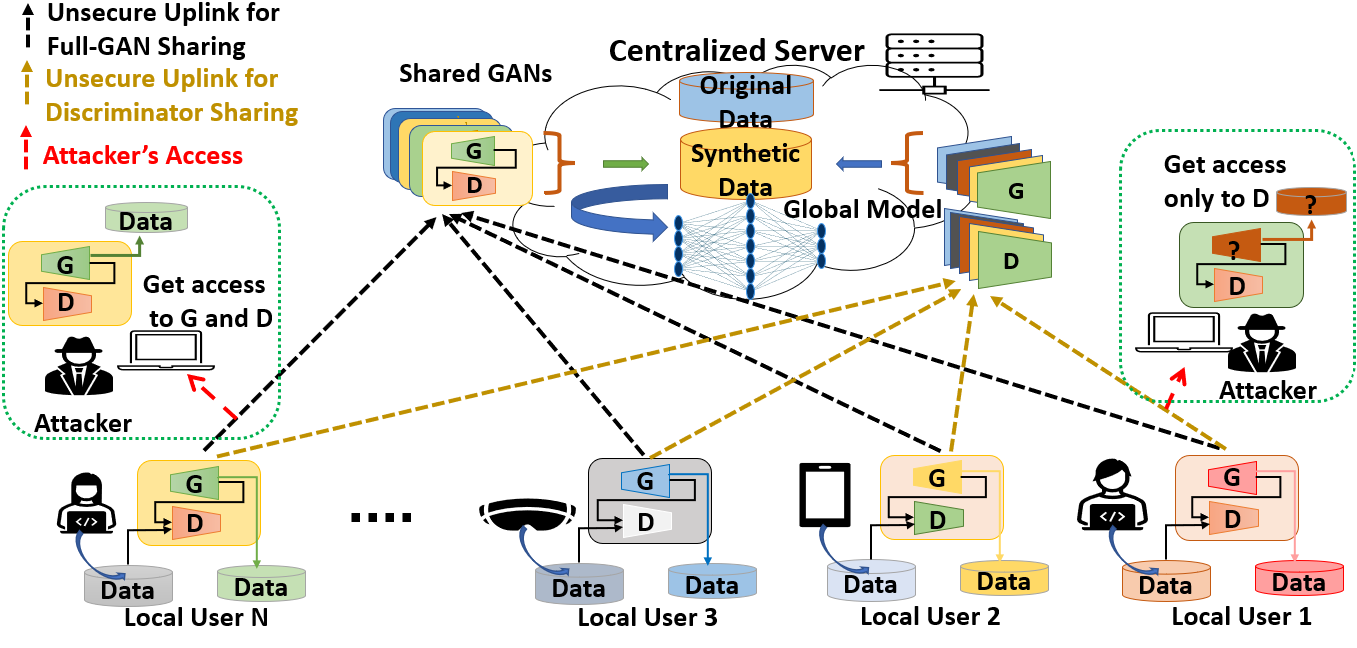}
  \caption{Comparison of full GAN sharing (left, marked in black lines) v.s. our proposed PS-FedGAN discriminator sharing (right, marked in yellow lines) in a scenario of $N$ users. In both cases, a local user trains its GAN model with a generator ($G$) and a discriminator ($D$). In traditional FL with fully shared GAN, full GAN models are 
  communicated, wheras PS-FedGAN only communicates discriminator to share. 
  In full GAN sharing, an attacker could eavesdrop on an unreliable channel to gain access to shared $G$ and $D$. As a result, the passive attacker can generate synthetic data which may approximate user data distribution. Furthermore, membership inference~\cite{membership} could also become
  viable in this setup ~\cite{logan}. In contrast, PS-FedGAN prevents eavesdropping
  of $G$ and denies attacker's access to both true GAN model $G$ and original data distributions.}
  \label{fig:1}
  \vspace*{-2mm}
\end{figure}

\textbf{Contributions: }In this work, we reexamine GAN sharing strategy in FL and propose a novel GAN publishing mechanism to address practical cases of non-IID data distribution among users. Through our proposed FL framework, named as \textbf{PS-FedGAN}, we reconstruct separate generators at the server from partially-shared GAN models trained locally at client users, in which each user only shares its discriminator with others.
The proposed PS-FedGAN significantly reduces the communication network overhead of model sharing and provides
better data privacy during communication rounds. Furthermore, it bridges the gap between utility and privacy. 
Figure~\ref{fig:1} highlights the distinction between existing full GAN sharing approaches 
and our proposed PS-FedGAN in an untrustworthy communication channel. 

We summarise our contributions as follows:

\begin{itemize}
    \item We propose a novel GAN-based FL learning framework, \textbf{PS-FedGAN}, to cope with non-IID data distributions among FL client users.
    More specially, we propose to train a generator at the server to capture the underlying data 
    distribution of the local user's GAN by only sharing a local discriminator. The proposed framework can significantly lower communication cost and improve data privacy.
    \item We provide analytical insights into the convergence of generator training
    at the client user end and at the cloud server. We investigate the convergence of
    the common discriminator training based on our proposed \textbf{PS-FedGAN}
    to establish the benefit of communication cost reduction by only sharing discriminators.
    \item We provide an interpretability analysis on privacy of the proposed \textbf{PS-FedGAN}, 
    which is further evaluated by our theoretical analysis and extensive experiments.
    \item We present experimental results against several well-known benchmark datasets to further demonstrate the efficacy and efficiency of our \textbf{PS-FedGAN}, in terms of utility, privacy, and communication 
    cost.

\end{itemize}

\section{Method and Architecture}
\label{method}
\subsection{Problem Setup}

In this work, we aim to develop a novel GAN-based FL framework for a global/common task in a distributed learning setup. For convenience, we will use image classification as an illustrative example henceforth.
Consider a reliable central server with limited access to client training data to achieve a desirable accuracy on the global task. Here, we assume non-IID data distributions and vulnerable communication channels among users, which are common in 
practical applications. For example, in smart healthcare,
one learning task could be to train neural networks for the detection of a specific disease.
One clinic may have a unique set of brain images from patients whereas other
neighboring hospitals may also have access to several types of similar diseases with a plethora of examples. A global model 
based on all distributed data 
to detect these diseases could benefit all hospitals and provide better service for patients. 
In such a collaborative system, local data
privacy is a critical concern. Also,
adding noises to the dataset 
to hide sensitive information may
lead to unwanted information distortion and/or artifacts.
In this work, we address the problem of how to preserve
local privacy while preserving the original data statistics. 

Motivated by existing GAN-based FL, we develop a novel GAN publishing mechanism for FL with privacy preservation and communication efficiency.
In alignment with other GAN-related FL works, we assume a 
system with a centralized server in the cloud
and multiple distributed clients/users in
communication with the server. Each client has access to enough resources to train a local GAN model. Note that, although we apply conditional GANs (cGAN) ~\cite{cGAN} to alleviate the need for label detection via pseudo-labeling, 
the principle of our proposed scheme generally applies to all types of generative models.

To assess the effectiveness of our framework in preserving privacy, we consider the presence of adversarial attackers. In order to model potential attacks from these adversaries, we make the assumption that an attacker has the capability to eavesdrop on the communication channels connecting local users and the server, as depicted in Figure~\ref{fig:1}. The attacker's objective is to estimate the data distribution of the local user through reconstruction attacks.

As illustrated in Figure~\ref{fig:1}, in conventional GAN-based FL methods, such an attacker would have unrestricted access to the entire GAN model. However, our proposed PS-FedGAN is specifically designed to address the security vulnerabilities associated with fully shared GAN models.

\subsection{PS-FedGAN}
We now delve into the structures of the proposed PS-FedGAN. In order to address the privacy concerns arising from full GAN sharing, we introduce a partial sharing approach within our PS-FedGAN framework. Specifically, we propose training two generators: generator $G_u$ at the local user's end and generator $G_s$ at the server side for each user.
In order to bridge the training of two generators, i.e.,
one at a local user and the other at the server
connected by a communication link, we share a common discriminator $D_u$ which is trained only at the local user. 

\begin{figure}[t]
  \centering
  \includegraphics[width=0.7\columnwidth]{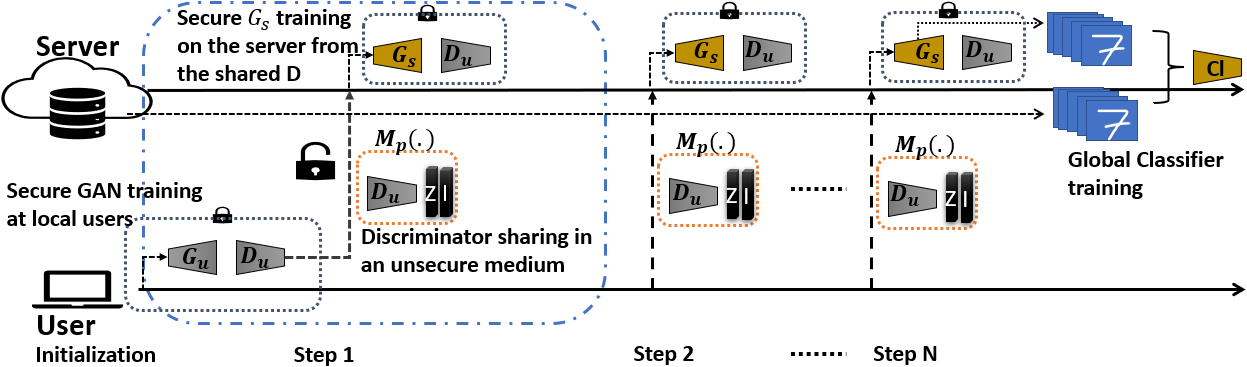}
  \caption{User-server communication process until one update of the global classifier ($C_l$)shown for one user using the PS-FedGAN method: The local user first starts training a local cGAN (generator $G_u$ and discriminator $D_u$). After a single batch training at the local user, trained $D_u$ is shared with the server using $\mathcal{M}_p$. $\mathcal{M}_p : \mathcal{M}(D_u,z,l)$, where $\mathcal{M}$ is a publishing mechanism. $z$ and $l$ are noise vectors and corresponding fake labels used to train $G_u$ at current step, respectively. At the server, we initiate a separate generator $G_s$ for each user. For every received user 
  update, we update $G_s$ by one step. After $N$ such steps, we update the $C_l$ using synthetic data generated from each $G_s$ and cloud available data. We continue this process till $C_l$ converges.}
  \label{fig:2}
  \vspace*{-2mm}
\end{figure}

To visually depict the update process of PS-FedGAN, we present the user-server communication flow leading up to a single-step update of the global model ($C_l$) in Figure~\ref{fig:2}.
In the case of a single user, the training process begins with the local user training a local GAN model consisting of generator $G_u$ and discriminator $D_u$. After completing a single batch training at the local user's end, the trained discriminator $D_u$ is shared with the server through the PS-FedGAN publishing mechanism $\mathcal{M}_p$. The details of this publishing scheme will be further elaborated in Section \ref{pub_scheme}. In our study,  we consider an untrustworthy communication channel where attackers may potentially gain access to the PS-FedGAN publishing mechanism $\mathcal{M}_p$. On the server side, we initialize a separate generator $G_s$ for each user, following the guidelines of PS-FedGAN Algorithm~\ref{alg:FedD}. Subsequently, the server updates the corresponding $G_s$ based on the information received through $\mathcal{M}_p$.

Before updating the global model $C_l$, we wait for a specific number of updates from the user side, denoted as $N$. In this context, $N$ can represent an epoch for the local user. Consequently, we carry out $N$ updates on both the local user's GAN and the corresponding $G_s$ in parallel. The generated data from all the updated $G_s$ models, corresponding to different local users, is combined with the server-available data to update the global model $C_l$. This cycle of updates and combination of generated data is repeated until $C_l$ converges to the desired point.

When deploying the FL models, we assume each local user and the server have agreed on a secret seed and the same architecture for $G_u$ and $G_s$. The secret seed serves
to initiate the weights of $G_u$ and its corresponding $G_s$. Note that each user is assigned its own dedicated generator in the cloud server. Also, we impose no constraint on $D_u$. 
PS-FedGAN follows a batch-wise training process where we update $G_s$, $G_u$, and $D_u$ at each step. Once $G_u$ and $G_s$ are initiated, local GAN training commences with $D_u$. In the following subsections, we provide a detailed explanation of the PS-FedGAN publishing mechanism $\mathcal{M}_p$ followed by an elaboration of the training process for both the local users and the cloud server in each communication round.

\subsubsection{PS-FedGAN Publishing Mechanism $\mathcal{M}_p$} \label{pub_scheme}

% \textcolor{red}{This section is not clear enough. Please first have a paragraph describing the general steps: we first train $G_u$ AND $D_u$ locally. Second, we apply $M_p$ to share data. Then, we do the server training with global classifier. After that, we transfer sth. back and do this again until converage. Organize your word and make it more clearly here. Also consider you could draw a figure to descripe this process clearly. For example, two lines parallel, one for server and one for client. From left to right represents time. Draw the connections between two lines to illustrate the whole process as Figure 2 above.}

In PS-FedGAN, let $z_t$ be the noise vector used to train $G_u$ with random labels $l_t$ at step/batch $t$ at the user side. Let $\mathcal{M}(\theta)$ be a user-side publishing mechanism used to send trained parameters $\theta$ to the server. The PS-FedGAN publishing mechanism is defined as $\mathcal{M}_p$: $\mathcal{M}(D_u,z_t,l_t)$, where the local user releases $D_u,z_t$, and $l_t$ to the server after each
training step of $D_u$. We shall show that
$\mathcal{M}_p$ preserves privacy and lowers communication cost compared to the full GAN releasing methods $\mathcal{M}(G_u, D_u)$ via theoretical analysis and experimental results in Section \ref{theory} and Section \ref{experiments}, respectively.

\subsubsection{PS-FedGAN Local User Training}

As discussed earlier, $D_u$ is trained at the local user 
end after initializing $G_u$. 
The local user has the flexibility to choose any architecture for $D_u$, as long as it is trainable with $G_u$. Following the conventional GAN training approach, $D_u$ is initially trained using real images and images generated by $G_u$ at each step. To handle the randomness of $G_u$ in a deterministic manner, the parameters of $D_u$ are updated during backpropagation (BP).

Depending on the convergence requirements of the architecture, multiple iterations of $D_u$ training can be performed. Subsequently, random vectors $z_t$ and corresponding random labels $l_t$ are generated for training $G_u$.  Before we train $G_u$, we release $\mathcal{M}_p$ to the server to 
minimize latency. Locally, $G_u$ is trained 
using $z_t$ and the corresponding $l_t$. Similar to general 
GAN training principles, the randomness in $D_u$ is handled 
deterministically, to enable the training of GAN $G_u$ accordingly. The parameters of $G_u$ are then updated through
BP.

\subsubsection{PS-FedGAN Server Training}

At the server, we maintain a dedicated generator ($G_s$) corresponding to each user initialized with the secret seed. 
Upon receiving the parameters $\mathcal{M}_p$ from the respective user, the training process for each $G_s$ starts. During this process, it is not necessary to wait for all other users to communicate with the server. Similar to the training of $G_u$, we train $G_s$ using the received $z_t$ and $l_t$ from the corresponding user. To ensure consistency and avoid randomness in $D_u$, we employ the same techniques used by the corresponding user during training.

Next, we perform BP through the parameters of $G_s$ and update them accordingly. Once we receive updates  from all users, we proceed to update the global classifier $C_l$. To update $C_l$ at each iteration, we generate a fixed number of samples from each $G_s$ and combine them with a portion of the available true data at the server. This approach allows us to create training samples that incorporate multiple user generators. 
We present the PS-FedGAN training algorithm in Algorithm~\ref{alg:FedD}

\begin{algorithm}
\caption{PS-FedGAN: Training Algorithm. Minibatch GAN training for distributed GANs. }\label{alg:FedD}
\begin{algorithmic}
% \Require Initial seed $s$ for weights, Generator model $\mathcal{G}$, and user id $i$
% \Ensure Initiate weights of user generator:$\mathcal{G}_u $ and cloud generator $\mathcal{G}_c $ with the same seed $s$. 
\For {each user $i$}
    \State Initial $G_{u_i}$ and $G_{s_i}$ using secret seed $key_i$
    \State Initiate the discriminator: $D_{u_i} $
\EndFor
\State Initiate the global model $C_l $ ( classifier ) and train using the available cloud data.

% \State $y \gets 1$
% \State $X \gets x$
% \State $N \gets n$
\For{each communication round, until $C_l$ is converged}
\For{number of training iterations}
%\While{Local GAN ($\mathcal{G}_u $,$\mathcal{D}_u $) converges}
\For{each step t}
 \State $z_t, l_t \gets$ random vectors, random labels
  \State Train $D_{u_i}$: $D_{u_i}^{(t+1)} \gets \mathcal{F}(G_{u_i}^{(t)}(z_l,l_t),\mathcal{D}_{u_i}^{(t)})$,  {where {$\mathcal{F}$ represent forward/back prop and weight updating }}
  \State Share $(\mathcal{D}_{u_i}^{(t+1)},z_t, l_t) : \mathcal{M}_p( \mathcal{D}_{u_i}^{(t+1)},z_t, l_t) $
  \State Train $G_{u_i}$: $G_{u_i}^{(t+1)} \gets \mathcal{F}(G_{u_i}^{(t)}(z_t, l_t), D_{u_i}^{(t+1)} )$
  
\State Train $G_{s_i}$: $G_{s_i}^{(t+1)} \gets \mathcal{F}(G_{s_i}^{(t)}(z_t, l_t), D_{u_i}^{(t+1)} )$
\EndFor
\EndFor
\State Train $C_l$ with cloud data and synthetic data : $C_l^{(t+1)} \gets {\mathcal{F}}(C_l^{(t)})$
\EndFor
% \label{alg:FED}
\end{algorithmic}
\end{algorithm}

%In this setup, we assume that the central server has no access to real data and there is no global model (i.e. classifier) at the server. 

%Since our focus is personalized FL we do not want to share down stream task with the server as well.

\subsection{Attacker Models} \label{attk}
To assess the performance of PS-FedGAN against potential attackers, we focus on reconstruction attacks specifically. These attackers are denoted as $\mathcal{A}_R$.
For any $\mathcal{M}(\theta)$, an attacker $\mathcal{A}_R$ attempts to reconstruct the training samples and assume $\mathcal{I}$ represents a reconstructed sample (images in this manuscript), i.e.,
\begin{equation}
    \mathcal{A}_R: \mathcal{M}(\theta) \mapsto \mathcal{I}.
\end{equation}
% where $\mathcal{I}$ represents a reconstructed sample (images in this manuscript).

In our experimental setups, we consider two types of attackers: $\mathcal{A}_1$ and $\mathcal{A}_2$. The bottleneck faced by any attacker against our model lies in 
the hidden generator model that is prone to information
leakage. Therefore, the process of reconstructing synthetic data helps preserve privacy in terms of the initial weights and network architecture, as further elaborated in Section \ref{theory}. Predicting the exact architecture has become challenging owing to advancements in deep learning techniques. 

For an attacker to obtain information, it is crucial to eavesdrop from the very beginning and ensure they do not miss any round of communication. Such requirements prove difficult for potential privacy attackers.
Additionally, the attacker must acquire knowledge of the initial weights of the generator, to be discussed in Section \ref{theory}. However, in practice, it is typically challenging, and often impossible, for an attacker to accurately estimate the generator's structure due to various practical constraints such as power, latency, and hardware
capabilities. The complexity and variability of generator architectures play a significant role in preserving privacy. The diverse range of architectures available for generators, coupled with their complexity, makes it highly challenging for an attacker to accurately infer the precise model structure. This inherent difficulty further enhances the preservation of privacy in the system. By utilizing complex and variable generator architectures, PS-FedGAN adds an additional layer of protection against potential privacy breaches.

Therefore, to define new attacks, we first introduce a factor $r \in [0,1]$. We assume that attackers $\mathcal{A}_1$ and $\mathcal{A}_2$ have the same architecture of $G_u$ and the same weights and bias terms of $G_u$ in every deep learning
network layer except the first. Let $w_1$ and $b_1$ be the layer 1 weight and bias term of $G_u$. Layer 1 weight $w_{a1}$ and bias term $b_{a11}$ of $\mathcal{A}_1$ are set to $w_{a11} = r w_1$ and $b_{a1} = b_1$, respectively. Similarly for $\mathcal{A}_2$, we set layer 1's weight unperturbed as $w_{a21} = w_1$, and set layer 1's bias to $b_{a21} = r b_1$. Clearly, these two attackers' knowledge of $G_u$ parameters
are only slightly different from the true parameters with
a multiplicative factor $r$ on Layer 1's weights and biases, respectively. 

% \begin{figure}
% \centering
% \includegraphics[width=1\columnwidth]{img/sys.jpg}
% \caption{Overview of the considered system and the proposed workflow}
% \label{f1}
% \end{figure}

\section{Theoretical Results}
\label{theory}

\vspace*{-2mm}

In this section, we provide a convergence analysis of the proposed method and provide insights into privacy preservation. Detailed corresponding proofs are presented in Appendix~\ref{apexB}.

\vspace*{-2mm}
\subsection{Convergence of $D_u$}
\vspace*{-0.5mm}
We first show the convergence of two generators and a discriminator trained according to \textbf{Algorithm 1}. Suppose that
the GAN trained at a local user consists of a generator $G_u$ capturing a probability distribution of $p_{gu}$, and a discriminator $D_u$.
Let the local user's data distribution be $p_{data}(x)$. The local user (client) trains $G_u$ with $z \sim p_z$. 
We have the following properties on model convergence.

\textbf{Proposition 1.} Two generators $G_u$ and $G_s$ trained on \textbf{Algorithm 1} with a shared discriminator $D_u$ converges to the same optimal discriminator $D_u^*$ as in~\cite{GAN1} and it uniquely corresponds to 
the given $G_u$, i.e.,
\vspace*{-2mm}
\begin{equation}
    D_u^* = \frac{p_{data}(x)}{p_{data}(x) + p_{gu}(x)}
\end{equation}

In Proposition 1, we see that the $D_u$ in PS-FedGAN converges to the same discriminator if we train $G_u$ and $D_u$ without $G_s$. That is, training $G_s$ in the cloud does not hamper the convergence or performance of 
the local GAN training. On the other hand, we have a unique $D_u$ given $G_s$. This property solidifies the convergence of $G_s$ to $G_u$ which is characterized in the following Propositions regarding the generator models.
% \textit{Proof} It is followed by the fact that \textbf{Algorithm 1} trains $D_u$ using $G_u$ and $G_s$ has no influence on $D_u$. Therefore, as in~\cite{GAN1} discriminator training is valid with the value function $V(G_u, D_u)$,

% \[V(G_u, D_u) = \int_{x} p_{data}\log(D_u(x)) + \int_{z} p_{z}(z)\log(1-D_u(G_u(z)))\,dz \]

% Using Radon-Nikodym Theorem, we have,
% \[E_{z\sim p_z(z)}\log(1-D_u(G_u(z))) = E_{x\sim p_{gu}(x)}\log(1-D_u(x))\]
% \[V(G_u, D_u) = \int_{x} p_{data}\log(D(x)) + p_{gu}(x)\log(1-D(x))\,dx \tag{1}\label{eq:1} \]

% Let $g(x) = a \log(x) + b \log(1-x)$ taking derivatives for stationary points: $\frac{\partial g(x)}{\partial x} = \frac{a}{x} - \frac{b}{1-x}$, which gives the unique maximizer $x = \frac{a}{a+b}$: which is indeed $D_u^*.$ The unique maximizer implies the uniqueness of $D_u^*$ for a given $G_u$.
\vspace*{-2mm}
\subsection{Convergence of $G_u$ and $G_s$}
\vspace*{-0.5mm}
\textbf{Proposition 2.} Two generators $G_u$ and $G_s$ trained according to \textbf{Algorithm 1} with a shared $D_u$ would converge to a unique $G^*= G_u^* = G_s^*$ which captures $p_{data}$.

This proposition establishes that two distributed generators trained using
PS-FedGAN converge to the same generative model. Moreover,
this model is the same as the optimized model that one can obtain via classical GAN training. Another vital observation is that both $G_u$ and $G_s$ capture the user-side data distribution.

\textbf{Proposition 3.} Any other generator $G_{\mathcal{A}}$ failing to capture the weights and architectures of $G_s$ or $G_u$, either in the initial state or in any single communication round, would fail to characterize the data distribution $p_{data}$. %\textcolor{red}{confirm this revised description correct or not}

This proposition provides us with insights into the capacity required by an attacker.
To attack the proposed model, an attacker would need to predict a generator $G$ using the information obtained from $\mathcal{M}_p$. However, to successfully carry out this attack, the attacker would need to possess precise knowledge of the generator's architecture, initial weights of either $G_s$ or $G_u$, and would need to monitor and capture every round of communication.  In the next section on experiments, numerical results shall substantiate the need for these requirements and further demonstrate the difficulty an attacker would face when attempting to breach the privacy of the PS-FedGAN model.

% \textit{Proof} It follows from the \textbf{proposition 1} and \textbf{proposition 2} that $p_{gu} = p_{data}$ and $p_{gs} = p_{data}$ satisfied only by a unique $G^*$ and $D_u^*$. As we have shown above, to obtain $G^* = G_u^* = G_s^*$ we need to have $ G_u^{'} =  G_s^{'}$ at each step, implies $G$ needs to capture all communication rounds. Therefore, these conditions restrict $G$ to capture $p_{data}$.

\vspace*{-2mm}
\subsection{DP property of $\mathcal{M}_p$}
We now discuss the DP properties of the proposed methods.

Let us denote the discriminator by $ D = f(data)$ and the generator by $G = g(D,z)$, 
where $z$ represents noise. From the post-processing property in~\cite{DP-1}, $g(f(D))$ is DP
if $f(data)$ is DP. Thus, the generator $G$ is DP given $D$ is DP. Furthermore, if the training process is based on original data, FL-GAN follows DP~\cite{DP-2}. 

In practical scenarios of PS-FedGAN,  accessed models of $D$ with quantized channel noise is DP ~\cite{AAAI-DP} on the original weights of $D$, i.e., $W_D$. Since an attacker can only gain access to discriminator $D$ during communication round in the proposed PS-FedGAN framework, any generators reconstructed by the attackers from the hacked $D$ is also DP on $W_D$.

\textbf{Proposition 4.} Any generator $G$ reconstructed by the attacker shall be DP in a communication
channel with quantization noise or channel induced error.

This proposition shows that the attackers' estimated GAN model is DP on the original model weights $W_D$ 
sent from a client to the server.
Considering the mutual information of shared models and original data, we have $I(data,W_D) \leq I(data,data)$. Therefore, if we preserve privacy in $W_D$, we also preserve some privacy in the original data.

\textbf{Proposition 5. }$\mathcal{M}_p$ preserves privacy compared to full data sharing.

\vspace*{-2mm}
\section{Experiments}
\label{experiments}
\vspace*{-1mm}
In this section, we present test results of the proposed PS-FedGAN under non-IID user data distributions. 
In the first subsection, 
we evaluate performance when utilizing the proposed method and provide performance comparison
with existing FL methods. We then present privacy measures and also evaluate the associated communication cost.  
Our experiments use several well-known benchmark datasets, including MNIST~\cite{deng2012mnist}, Fashion MNIST~\cite{fashionMnist}, SVHN~\cite{svhn}, and CIFAR10~\cite{cifar10}.

\vspace*{-2mm}
\subsection{Evaluation of Utility}
\vspace*{-0.5mm}

Our study considers three distinct scenarios of heterogenous user cases, which are based on the work presented in
~\cite{IJCAL}. Each case involves a total of 10 users and
the following details: 

\begin{itemize}
    \item[Split-1:] In this case,  training data is divided into 10 shards, each containing samples from a single class. Each user is randomly assigned one distinct shard.
\item[Split-2:]  This case generates 20 training data shards, each consisting of samples from a single class. Two shards are randomly assigned to each user without overlap.
\item[Split-3:] This case generates 30 shards, each containing samples from a single class. Three shards are randomly assigned to each user without overlap.\end{itemize}

As a utility measure, we select the classification accuracy of the global model in a supervised setup. We compare our results against several existing FL alternatives: FedAvg (FA) ~\cite{Fed1},
FedProx (FP) ~\cite{fedprox},
SCAFFOLD (SD) ~\cite{scaffold},
Naivemix (NM) ~\cite{yoon2021fedmix},
FedMix (FM)~\cite{yoon2021fedmix}, and
SDA-FL (SL)~\cite{IJCAL} as baselines. We also compare our partially-shared PS-FedGAN (PS) with the fully-shared GAN (FG)
as a performance benchmark.
The results of Split-1/2 are shown in Table \ref{tab:1}.
The performance of Split-3 can be found in the Appendix.

Table~\ref{tab:1} shows the superior performance of the proposed PS-FedGAN over most existing methods except for 
the FG. More specially, we see a significant improvement in PS-FedGAN in Split-1 for the CIFAR10 dataset. 
Compared with full-GAN (FG) sharing, our partially-shared GAN
achieves similar performance but at significantly reduced communication cost and privacy loss as further shown in Table \ref{com1}. 
Note that in the above test, the SDA-FL method requires an infinite privacy budget to achieve the desired utility. This indicates that SDA-FL faces significant challenges in balancing privacy and utility.
Additionally, it is worth mentioning that other alternatives provide more real data for the classifier from Split-1 to Split-3, whereas in our proposed method, we maintain a constant amount of real data on the server (1\%). Similar results can be seen for Split-3 in the Appendix.

\begin{table}[htb]
\centering
\caption{Classification accuracies (best) of existing FL methods and full GAN sharing compared with PS-FedGAN in Split-1 and Split-2. Some of the results are from~\cite{IJCAL}. We assume 1\% of the training data of each dataset is available in the cloud for FG and PS-FedGAN. } 
%We use the following abbreviations. FedAvg: FA, FedProx: FP, SCAFFOLD: SD, Naivemix: NM, FedMix: FM, SDA-FL: SL, PS-FedGAN: PS, Sharing full cGAN: FG  }

\begin{tabular}{lll l l llll}
\toprule
\multicolumn{1}{c}{}&\multicolumn{4}{c}{Split-1}&\multicolumn{4}{c}{Split-2}\\
\cmidrule(r){2-5}\cmidrule(r){6-9}
 % \multicolumn{1}{c}{Split }     & 1 & 2 & 3            \\
 %    \cmidrule(r){2-5}
Method & MNIST  & FMNIST & SVHN  & CIFAR10 & MNIST  & FMNIST & SVHN  & CIFAR10\\ 
\midrule
FA &  83.44 & 16.50 & 14.05 & 18.36 & 97.61 & 73.50 & 81.11 & 61.28\\ 
FP & 84.17 & 57.14 &17.53 & 11.24 & 97.55 & 75.76 &86.28 &63.16\\ 
SD & 25.39 & 56.80 &11.64 & 12.81 & 94.17 & 70.82 &73.34 & 60.78\\ 
NM & 84.35 & 66.62 &14.35 & 14.39 &  84.35 & 79.54 &84.64 & 64.39\\ 
FM & 90.96 & 72.11& 16.78 & 13.57 & 90.96 & 82.41& 86.61 & 65.76\\ 
SL & 98.19 & 85.70 & 88.46 & 37.70 & 98.26 & 86.87 & 90.70 & 67.89\\ 
FG & 98.41 & 88.77& 91.92& 66.98 & 98.41 & 89.01 & 92.61 & 69.91\\
\midrule
{PS} & {98.31} & {88.42} & {91.73} & {66.96} & {98.44} & {89.02} & {92.30} & {69.89}\\ 
 \bottomrule
\end{tabular}
\label{tab:1}
\end{table}

% \begin{table}[h]
% \centering
% \caption{Classification accuracies of existing FL methods and PS-FedGAN in split-2. We assume 1\% of the training data of each dataset is available at the cloud}
% \begin{tabular}{lll l l}
% \toprule
%  % \multicolumn{1}{c}{Split }     & 1 & 2 & 3            \\
%  %    \cmidrule(r){2-5}
% Method & MNIST  & FMNIST & SVHN  & CIFAR10\\ 
% \midrule
% FedAvg &  97.61 & 73.50 & 81.11 & 61.28\\ 
% FedProx & 97.55 & 75.76 &86.28 &63.16 \\ 
% SCAFFOLD & 94.17 & 70.82 &73.34 & 60.78 \\ 
% Naivemix & 84.35 & 79.54 &84.64 & 64.39\\ 
% FedMix & 90.96 & 82.41& 86.61 & 65.76\\ 
% SDA-FL & 98.26 & 86.87 & 90.70 & 67.89 \\ 
% \textbf{PS-FedGAN} & \textbf{98.44} & \textbf{89.02} & \textbf{90.99} & \textbf{69.89} \\ 
% Sharing full cGAN & 98.41 & \\
%  \bottomrule
% \end{tabular}
% \label{cl3}
% \end{table}

\subsection{Privacy Evaluation}

We now evaluate the privacy of PS-FedGAN with respect to reconstruction attacks, which is often viewed as one of the most dangerous attacks. Here, we utilize the MNIST dataset.
To evaluate the efficacy against reconstruction attacks, we use several proxies to measure the privacy leakage, including normalized mean square error (NMSE), structural similarity index (SSIM) ~\cite{ssim}, and classification accuracy. We consider two different non-IID user setups:
\vspace*{-1mm}
\textcolor{black}{
    \begin{itemize}
        \item Setup-1 includes three users. User-1 has access to classes \{0,1\}, user-2 has access to classes \{2,3,4\}, and user-3 has access to remaining classes \{5,6,7,8,9\}. Also attacker model $\mathcal{A}_{1}$ described in Section \ref{attk} is used.
        \item In Setup-2, we consider 10 users with data splitting in Split-1, where attacker model $\mathcal{A}_2$ is applied here in reconstruction attacks. 
    \end{itemize}}
    \vspace*{-1mm}

\textbf{Classification Accuracy:} To evaluate the attacker's classification accuracy on reconstructed images, we first consider Setup-1. In this setup, we assume that the attacker has access to all the elements released by the releasing mechanism $\mathcal{M}p$. Furthermore, we assume that the attacker can accurately guess the exact generator architecture. Note that at the beginning of training, the generator weights are different for $\mathcal{A}_{1}$ ($w{a11} = r w_1$), from the cloud generator as mentioned earlier.
The generators are trained
simultaneously at the user, the server, and the attacker. 

We then pick an attacking classifier trained on the original MNIST training set and infer data generated by the attacker's generator. Here, the attacker assumes 
that the user works with MNIST data. Table~\ref{t1} illustrates the attacker's performance in different $r$. From Table~\ref{t1}, we see that the attacking gains
better accuracy with increasing $r$. In Table~\ref{t1}, we evaluate the classification accuracy at the cloud using the same attacking classifier, which reflects the potential of an
ideal attacker and the utility at the cloud. Table~\ref{t1} shows that an attacker with some disparity in the initial weights can only perform like a random guess. Note that in this scenario, user-1 has 2 classes, user-2 has 3 classes, and user-3 has 5 classes. These results suggest that an attacker must obtain very accurate information on architecture and initial model weights to achieve higher
and nontrivial inference accuracy. Since such requirements
are improbable and impractical, our results 
establish the robustness of our proposed method against classification attacks.

\begin{table}[h]
\centering
\caption{Classification accuracies (best) on attacker's generator and cloud's generator: How attacker's performance (classification accuracy \%) varies as $r$ varies on $\mathcal{A}_1$ on Setup-1}
\begin{tabular}{lllllll}
\toprule
\multicolumn{1}{c}{}&\multicolumn{3}{c}{Attacker's Models}&\multicolumn{3}{c}{Cloud's models}                 \\
    \cmidrule(r){2-4}\cmidrule(r){5-7}
r(on attacker) &  User1 &  User2 & User3 &  User1 &  User2 & User3\\ 
\midrule
$1-1\times10^{-4}$ &  0.4955 & 0.3390 & 0.2048 &  0.9911 & 0.9902 & 0.9722\\ 
$1-1\times10^{-7}$ &  0.4933 & 0.3323 & 0.1945 &  0.9983 & 0.9727 & 0.9610\\ 
$1-1\times10^{-15}$ &  0.4941 & 0.3299 & 0.1990  &  0.9980 & 0.9677 & 0.9690 \\ 
$1-1\times10^{-21}$ &  0.5027 & 0.3341 & 0.2056  &  0.9981 & 0.9746 & 0.9754  \\ 
\bottomrule
\end{tabular}
\label{t1}
\end{table}

% \begin{table}[h]
% \centering
% \caption{The Clouds performance (\%) as attacker\'s initial weights (layer 1) ratio ($\frac{predicted W}{True W}$) varies on 3 different users case}
% \begin{tabular}{llll}
% \toprule
% r (on attacker) &  User1 &  User2 & User3\\ 
% \midrule
% $1-10^{-4}$  \\ 
% $1-10^{-7}$ \\ 
% $1-10^{-15}$ \\ 
% $1-10^{-21}$\\ 
% \bottomrule

% \end{tabular}
% \label{t2}
% \end{table}
\begin{figure}
  \centering
  \includegraphics[width=0.6\columnwidth]{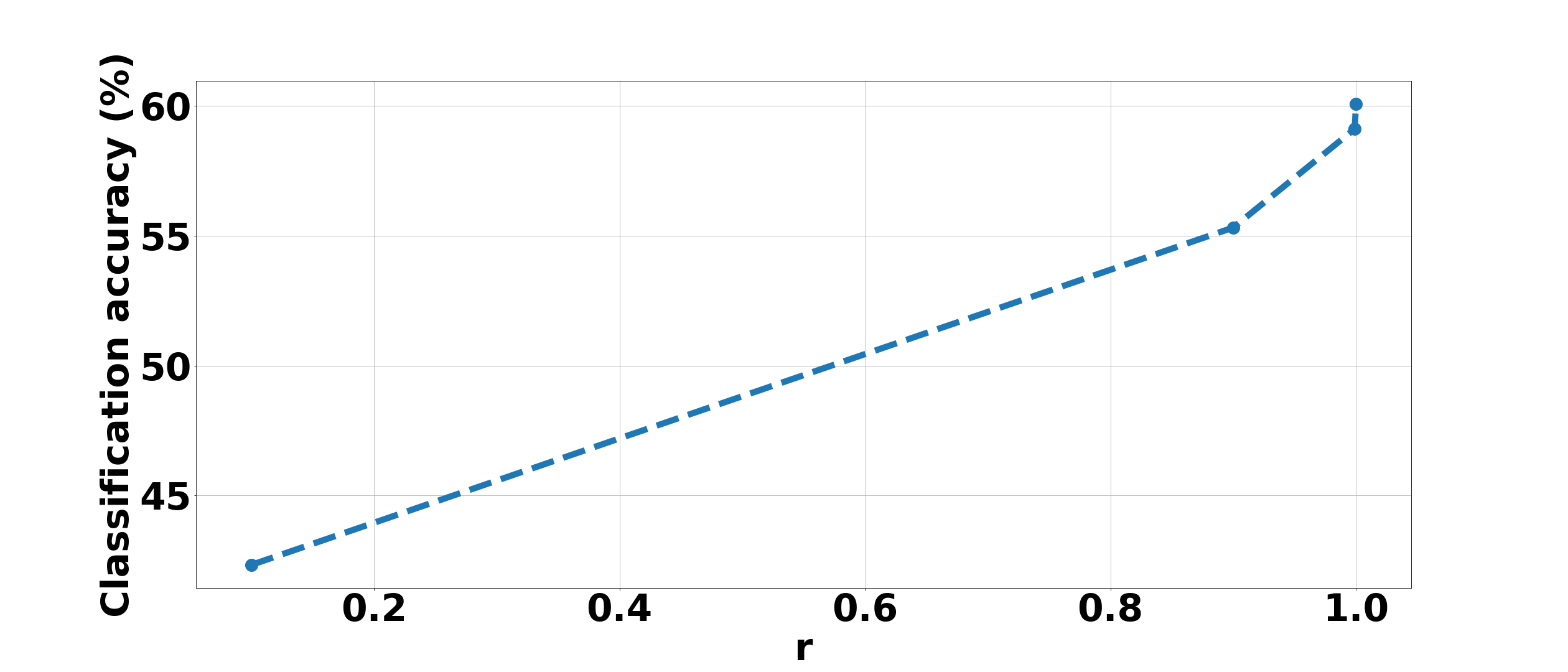}
  \vspace*{-1mm}
  \caption{Reconstruction attacks: Different $r$ from 0.1 to 0.999999 evaluated under $\mathcal{A}_2$.  }
  \label{fig:3}
\vspace*{-2mm}
\end{figure}

Next, we consider $\mathcal{A}_2$'s classification accuracy on reconstructed images. For this
attacker, the difference between its generator and the cloud generator lies the first layer bias term ($b_{a21} = r b_1$).
% At the begging of the training, we set the attacker's first layer bias $= r \times$ user's first layer bias.
We evaluate $\mathcal{A}_2$'s performances on Split-1.  Figure~\ref{fig:3} illustrates the attacker's performance for different $r$. Figure~\ref{fig:3} shows that the attacking gains
better classification with increasing $r$. 
Similar to the first type of attackers in $\mathcal{A}_1$,
these results suggest that an attacker from $\mathcal{A}_2$
also needs to have very accurate architecture 
knowledge and the initial weights. This test case further establishes the 
robustness of PS-FedGAN  robust against classification attacks.

% \begin{table}[h]
% \centering
% \caption{Reconstruction attacks: How attacker\'s performance varies as the estimated initial weights (bias of the layer 1) ratio ($\frac{predicted W}{True W}$) varies}
% \begin{tabular}{llll}
% \toprule
% r & Attacker\'s classification accuracy (\%)\\
% \midrule
% 0.1 &  42.33\\ 
% 0.9 & 55.32\\ 
% 0.999 & 59.12  \\ 
% 0.999999 & 60.09 \\ 
% \bottomrule
% \end{tabular}
% \label{t3}
% \end{table}

\textbf{Reconstruction Quality and Similarity}   As presented in the paper~\cite{BMCV}, reconstruction quality and similarity can be used as a measure of privacy leakage. For this, we consider 2 metrics, i.e., SSIM and NMSE, in Setup-1.  Table~\ref{t4}  compares the corresponding generations of the attacker and the cloud based on the similarity between each generated image. From Table~\ref{t4}, we can see that the NMSE values achieved by the attackers are high while SSIM (maximum 1) is very low. These results indicate  that the attacker-generated images cannot capture the true data distribution. We shall further show in Appendix that, if the attacker deviates from the actual generator weights, its convergence would be to a trivial point. As a result, no underlying user data properties are captured, as illustrated by the visual examples in Figure~\ref{fig:4}. More results and discussions are provided in the Appendix.

\begin{table}[h]
\centering
\caption{Reconstruction attacks: Attacker $\mathcal{A}_1$ reconstruction quality on 3 different users in Setup-1. For NMSE lower values are better and for SSIM higher values are better.}
\begin{tabular}{llll}
\toprule
Metric &  User1 &  User2 & User3\\ 
\midrule
NMSE &  1.4108 & 1.4585 & 1.2553 \\ 
SSIM &  0.01956 & 0.0253 & 0.0229\\ 
\bottomrule

\end{tabular}
\label{t4}
\vspace*{-3mm}
\end{table}

\begin{figure}[htbp]
  \centering
\subfigure[Regenerated data samples in the cloud server.]{ \includegraphics[height=0.15\columnwidth,width=0.25\columnwidth]{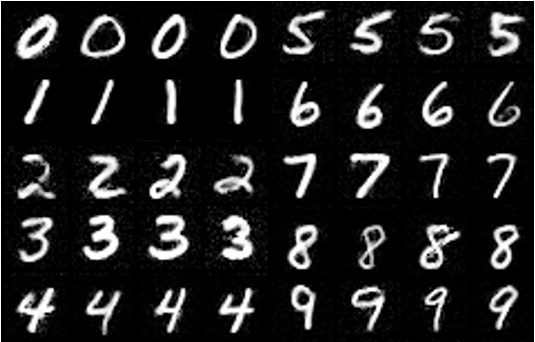}}
\hfill
  \subfigure[Regenerated samples of $\mathcal{A}_2$ (each \# for one user). ]{ \includegraphics[height=0.15\columnwidth,width=0.25\columnwidth]{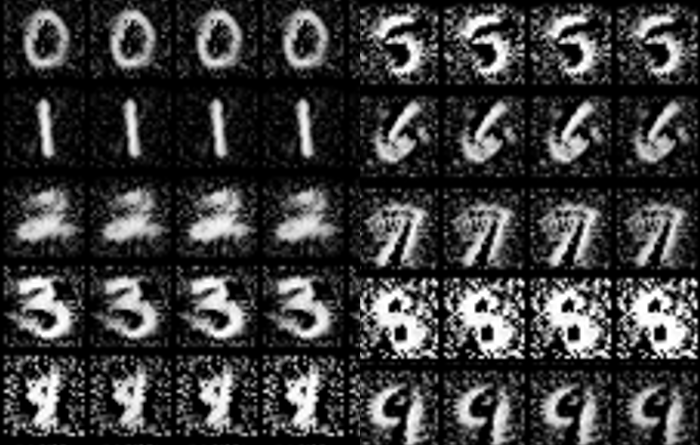}}
  \hfill
\subfigure[Regenerated samples of $\mathcal{A}_1$ (each block for one user). ]{ \includegraphics[height=0.15\columnwidth,width=0.25\columnwidth]{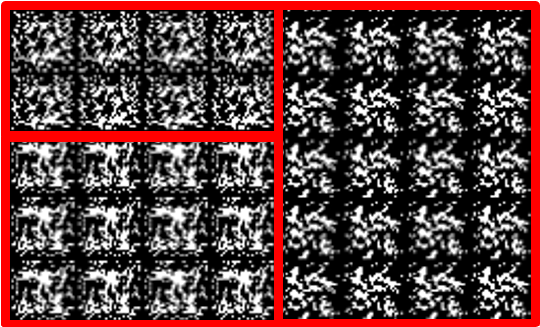}}
% \subfigure[Loss curves of parallel training of cloud generator $G_s$ and the generator of $\mathcal{A}_1$: $G_A$ and user discriminator $D_u$]{ \includegraphics[width=0.6\columnwidth]{images/Fig2.png}}
\vspace*{-2mm}
  \caption{ Divergence of the attacker. Any attacker fails to initiate with the same parameters as $G_u$ or $G_s$ would fail to capture the local user's distribution ($r=0.9999$ for both $\mathcal{A}_1$ and $\mathcal{A}_2$). }
  \label{fig:4}
\end{figure}

\subsection{Evaluation of communication Cost}
\vspace*{-2mm}

\begin{table}[htbp]
\centering
\caption{Number of parameters communicated at each step for Split-3 }
\begin{tabular}{lll }
\toprule
 % \multicolumn{1}{c}{Split }     & 1 & 2 & 3            \\
 %    \cmidrule(r){2-5}
Dataset & Full GAN sharing: $\mathcal{M}(G,D)$  & PS-FedGAN: $\mathcal{M}_p(D,z,l)$ \\ 
\midrule
MNIST &  3 M & 1.5 M \\ %(3002915) & 1496074+12120 \\ 
FMNIST & 5 M & 1.5 M \\%4945092 &  1413697+12928\\ 
SVHN & 2 M &  0.7 M \\%1797843 & 667457+ 28416 \\ 
CIFAR10& 6.8 M & 2.98 M \\ 

 \bottomrule
\end{tabular}
\label{com1}
\end{table}

Table~\ref{com1} illustrates the number of parameters that need to be shared between each user and the cloud server per communication round in the full GAN vs PS-FedGAN. From the table, we can see a significant saving in PS-FedGAN compared to classical full GAN sharing. The advantage applies similarly to various different GAN architectures.

\vspace*{-2mm}
\section{Conclusion and Future Works}
\vspace*{-2mm}
In this work, we develop a GAN-based FL framework PS-FedGAN, which can beneficially preserve local data privacy and reduce communication overhead in comparison with  existing FL proposals. Our novel PS-FedGAN achieves 
learning utility comparable to that of fully shared GAN architecture, while significantly securing data privacy and lowering the communication cost.
Empirical results further demonstrate superior results against state-of-the-art GAN-based FL frameworks. This 
PS-FedGAN principle and architecture can be directly generalized 
to incorporate any existing GAN. 
In future work, we plan further explore the effect
of lossy networking channels and improve PS-FedGAN's robustness against
non-ideal network links.
Another promising extension of this work is to explore the model and data heterogeneity presented in~\cite{future1}. 

\newpage
%%%%%%%%%%%%%%%%%%%%%%%%%%%%%%%%%%%%%%%%%%%%%%%%%%%%%%%%%%%%
\bibliographystyle{plainnat}
\bibliography{ms}
\newpage

\section*{Appendix }
\label{apex1}

\section*{A Backgrounds and Related Works}
\label{RW}

\subsection*{A 1.1 Classical Federated Learning}

Federated Learning (FL) offers a 
simple and efficient method for privacy-protected distributed learning. In classic FL, local data collections are often non-IID, unbalanced, and widely distributed, typically with limited communication bandwidth~\cite{Fed1}. To reduce communication overhead and maintain privacy, a Federated Average (FedAvg) algorithm is proposed. FedAvg aggregates gradient updates from different users by combining several local updates in a single round of communication~\cite{FL}.

The convergence of FedAvg for non-IID data is discussed in~\cite{ICLR}. However, as shown in~\cite{fednon}, the accuracy loss of FedAvg can be up to $\sim 55\%$ in a non-IID setup for some datasets. How to effectively handle non-IID data distributions among different users remains an open FL question. First, FedProx~\cite{fedprox} is proposed
as a generalized FedAvg for heterogeneous networks. 
SCAFFOLD is another extension of FedAvg  that incorporates predictive variance reduction techniques~\cite{scaffold}. In~\cite{NEURIPS2021_2f2b2656}, the authors propose a classifier calibration with virtual representation, 
leveraging a Gaussian Mixer Model to sample virtual features.  As an alternative solution, the authors of ~\cite{Rawdata} proposed raw data sharing in a cooperative learning mechanism. In ~\cite{ICLR}, the authors proposed 
the use of some initial training data e.g., 5\%, to handle non-IID data with traditional FL algorithms.
Some other classic methods for non-IID data include GAN sharing ~\cite{TNNLS,IJCAL,PerFEDGAN}, synthetic data sharing ~\cite{syndata,syndata2} and global sub-dataset sharing~\cite{fednon}.

% $\cdots$. \textcolor{red}{include more related literature on non-IID data distribution, group similar works to the same category and list their citation number together without detailed discussion, especially published in previous Neurips}.

\subsection*{A 1.2 GAN-based FL for IID and Non-IID Clients}
Training GANs locally before sharing trained models with a centralized server is a
popular approach for handling both IID and non-IID data distributions. 
In~\cite{IJACI-2}, the authors applied a conditional GAN (cGAN) and shared
local classifier and generator with the central server which trains a global classifier and a generator to guide
local users. Applying similar concept, the authors of \cite{ijacal-3} proposed to incorporate model splitting to keep part of the cGAN (discriminator) and part of the hidden classifier by sharing the generator and a global classifier. In addition, as introduced in~\cite{IJCAL,PerFEDGAN}, one can share the entire local GAN with the server and create a synthetic dataset to train a global GAN. Similarly, full GAN sharing has been discussed in ~\cite{TNNLS}. Only shared generators with the maximum mean discrepancy are aggregated. To address  privacy concerns, DP~\cite{DP1} has been applied in GAN-based FL \cite{DP-GAN, IJCAL}, while a privacy budget is introduced to balance privacy and utility.

\section*{B Theoretical Analysis }
\label{apexB}
We provide convergence guarantees of the proposed method and provide insights into why attacking is hard. First, to observe the convergence of distributed two generators with a shared discriminator trained on proposed \textbf{Algorithm 1}, consider a GAN at the local user side with a generator $G_u$ capturing a probability distribution of $p_{gu}$ with the corresponding discriminator $D_u$ and the dedicated generator $G_s$ at the server side. Assume user's data distribution as $p_{data}(x)$ and the local user trains $G_u$ with $z \sim p_z$
\subsection*{B 1.1 Convergence of $D_u$}
\textbf{Proposition 1.}  Two generators $G_u$ and $G_s$ trained on \textbf{Algorithm 1} with a shared discriminator $D_u$ converges to the same optimal discriminator $D_u^*$ as in~\cite{GAN1} and it uniquely corresponds to 
the given $G_u$, i.e.,
\begin{equation}
    D_u^* = \frac{p_{data}(x)}{p_{data}(x) + p_{gu}(x)}.
\end{equation}

\textit{Proof}: In the proposed PS-FedGAN, \textbf{Algorithm 1} trains $D_u$ using $G_u$ while $G_s$ has no influence on $D_u$. Therefore, as illustrated in~\cite{GAN1}, the discriminator training is valid with the value function $V(G_u, D_u)$ denoted by
\begin{equation}
    V(G_u, D_u) = \int_{x} p_{data}\log(D_u(x))\,dx + \int_{z} p_{z}(z)\log(1-D_u(G_u(z)))\,dz.
\end{equation}

Using the Radon-Nikodym theorem, we have the following conclusions:
\begin{equation}
    E_{z\sim p_z(z)}\log(1-D_u(G_u(z))) = E_{x\sim p_{gu}(x)}\log(1-D_u(x)).
    \end{equation}
    Then, the value function can be recalculated as
\begin{equation}\label{eq:1}
V(G_u, D_u) = \int_{x} p_{data}\log(D(x)) + p_{gu}(x)\log(1-D(x))\,dx.
\end{equation}

Let $g(x) = a \log(x) + b \log(1-x)$. The derivatives for stationary points can be calculated as
\begin{equation}
    \frac{\partial g(x)}{\partial x} = \frac{a}{x} - \frac{b}{1-x},
\end{equation}
and
\begin{equation}
   \frac{\partial^2 g(x)}{\partial x^2} = -\frac{a}{x^2} - \frac{b}{(1-x)^2},
\end{equation}

which give the unique maximizer $x = \frac{a}{a+b}$ with $\frac{\partial^2 g(x)}{\partial x^2} < 0$ for $a,b \in (0,1)$.

Hence, the optimal $D_u^*$ can be calculated by
\begin{equation}
    D_u^* = \frac{p_{data}(x)}{p_{data}(x) + p_{gu}(x)}.
\end{equation}

Here, the unique maximizer implies the uniqueness of $D_u^*$  for a given $G_u$.

\subsection*{B 1.2 Convergence of $G_u$ and $G_s$}
\textbf{Proposition 2.} Two generators $G_u$ and $G_s$ trained according to \textbf{Algorithm 1} with a shared $D_u$ would converge to a unique $G^*= G_u^* = G_s^*$ which captures $p_{data}$.

\textit{Proof}: Following \textbf{Proposition 1}, \textbf{Algorithm 1} converges to $D_u^*$. Therefore, in the training of the generator, we have the following optimization formulations, i.e.,

\begin{equation}\label{eq:2}
    G_u^* = \argmin_{G_u}V(G_u,D_u^*),
\end{equation}
and
\begin{equation} \label{eq:3}
    G_s^* = \argmin_{G_s}V(G_s,D_u^*).
\end{equation}

According to \textbf{Algorithm 1}, after each epoch of training in $G_u$  and $G_s$, we have $ G_u^{'} =  G_s^{'}$. This is because the input to both networks and initial weights are the same, leading to the same loss with the same gradients to be updated in the backpropagation.
 %since we handle randomness in a deterministic way. 
 Therefore, the training of $G_s$ can be viewed as the traditional GAN training with $G_s$ and $D_u$. Since $z$ is shared in the communication, we have
 \begin{equation}\label{eq:4}
     V(G_s, D_u) = \int_{x} p_{data}(x)\log(D_u(x))\,dx + \int_{z} p_{z}(z)\log(1-D_u(G_s(z)))\,dz.
 \end{equation}
 Moreover, if $G_s$ captures a distribution $p_{gs}$, from Eq. (\ref{eq:4}), we have
\begin{equation}
    V(G_s, D_u) = \int_{x} p_{data}(x)\log(D_u(x))\,dx + \int_{z} p_{z}(z)\log(1-D_u(G_u(z)))\,dz,
\end{equation}
 which could be further calculated via
\begin{equation}\label{eq:5} 
    V(G_s = G_u, D_u) = \int_{x} p_{data}(x)\log(D(x)) + p_{gs}(x)\log(1-D(x))\,dx. 
\end{equation}

The uniqueness of $D_u^*$ in Eq. (\ref{eq:1}) and Eq. (\ref{eq:5}) leads to
$p_{gu} = p_{gs}$. Thus, $G_s$ captures the same distribution as $G_u$. Hence, we can train two generators to capture the same distribution by only sharing a discriminator, without sharing original data explicitly. If $G_u$ achieves $G_u^*$, we have $G_u^* = G_s^*$. Then the optimization problems in Eq. (\ref{eq:2}) and 
Eq.~(\ref{eq:3}) reduce to
\begin{equation}
    G^* = G_u^* = G_s^* = \argmin_{G_u}V(G_u,D_u^*).
\end{equation}

In order to prove that $G_s^*$ captures $p_{data}$, we refer to~\cite {GAN1}. For $G^*$, from the viewpoint of game theory, $D_u$ fails to distinguish between true and fake data. Then, we have
\begin{equation}
    D_u^* = \frac{p_{data}}{p_{data} + p_{gu}} = \frac{1}{2},
\end{equation}
which leads to $p_{gu} = p_{data}$. Now, we have $p_{gs} = p_{data}$. Thus, the server generator also captures the data distribution of the corresponding local user. Hence, 
$G^*$ captures $p_{data}$.
The uniqueness of $G^*$ shall follow the same conclusion from~\cite{GAN1}. To prove this, we first assume that $p_{gu} = p_{data}$. Then the value in Eq.  (\ref{eq:5}) can be calculated as
\begin{align}
    V(G, D_u^*) &= \int_{x} p_{data}(x)\log(D_u^*) + p_{gs}(x)\log(1-D_u^*)\,dx\\
     &=\int_{x} p_{data}(x)\log(0.5) + p_{gs}(x)\log(0.5)\,dx\\
     &= \log(0.5)(\int_{x} p_{data}(x) + p_{gs}(x))\,dx  \\
     &=-\log(\frac{1}{4}).
\end{align}

%\[V(G, D_u^*) = \int_{x} p_{data}(x)\log(D_u^*) + p_{gs}(x)\log(1-D_u^*)\,dx  \]

%\[V(G, D_u^*) = \int_{x} p_{data}(x)\log(0.5) + p_{gs}(x)\log(0.5)\,dx  \]

%\[V(G, D_u^*) = \log(0.5)(\int_{x} p_{data}(x) + p_{gs}(x))\,dx  \]
%\[V(G, D_u^*) = -\log(\frac{1}{4})\]

This provides us with the global minimum.
On the other hand, for any $G$ and $D_u^*$, Let $M(G) = \max_D V(G,D)$. Then, we have,
% \[\max_D V(G,D) = \int_{x} p_{data}(x)\log(D_u) + p_{gs}(x)\log(1-D_u)\,dx \]
\begin{align}
 M(G) &=
\max_D V(G,D) \\&= \int_{x} p_{data}(x)\log(D_u) + p_{gs}(x)\log(1-D_u)\,dx \\
   &= \int_{x} p_{data}(x)\log(\frac{p_{data}}{p_{data} + p_{gu}}) + p_{gs}(x)\log(1-\frac{p_{data}}{p_{data} + p_{gu}})\,dx\\&=\int_{x} p_{data}(x)\log(\frac{p_{data}}{p_{data} + p_{gu}}) + p_{gs}(x)\log(\frac{p_{gu}}{p_{data} + p_{gu}})\,dx.
\end{align}
%\[M(G) = \int_{x} p_{data}(x)\log(\frac{p_{data}}{p_{data} + p_{gu}}) + p_{gs}(x)\log(1-\frac{p_{data}}{p_{data} + p_{gu}})\,dx \]
%\[M(G) = \int_{x} p_{data}(x)\log(\frac{p_{data}}{p_{data} + p_{gu}}) + p_{gs}(x)\log(\frac{p_{gu}}{p_{data} + p_{gu}})\,dx \]

After some manipulations, $M(G)$ can be calculated as
\begin{align}
    M(G) &= -\log(4)+ \int_{x} p_{data}(x)\log(\frac{p_{data}}{\frac{(p_{data} + p_{gu})}{2}}) + p_{gs}(x)\log(\frac{p_{gu}}{\frac{(p_{data} + p_{gu})}{2}})\,dx\\
    &=-\log(4)+ 2JSD(p_{data}|p_G),
\end{align}

%\[M(G) = -\log(4)+ \int_{x} p_{data}(x)\log(\frac{p_{data}}{\frac{(p_{data} + p_{gu})}{2}}) + p_{gs}(x)\log(\frac{p_{gu}}{\frac{(p_{data} + p_{gu})}{2}})\,dx \]

%\[M(G) = -\log(4)+ 2JSD(p_{data}|p_G),\]

where JSD is the Jenson-Shannon Divergence which is non-negative. This yields to $-\log(4)$ which is the global minimum. Finally, it gives us $p_{data} = p_{gu}$, and the uniqueness of $G^*$ is proved.
\subsection*{B 1.3 Divergence of any $G$ other than $G_u$ or $G_s$}
\label{ss:div}

\textbf{Proposition 3.} Any other generator $G_{\mathcal{A}}$ failing to capture the weights and architectures of $G_s$ or $G_u$, either in the initial state or in any single communication round, would fail to characterize the data distribution $p_{data}$. 

\textit{Proof}: From \textbf{Proposition 1} and \textbf{Proposition 2}, we have $p_{gu} = p_{data}$ and $p_{gs} = p_{data}$ which are made possible only by a unique $G^*$ and $D_u^*$. As shown before, to obtain $G^* = G_u^* = G_s^*$, we need to have $ G_u^{'} =  G_s^{'}$ at each step, implying that $G$ needs to capture all communication rounds. Therefore, these conditions restrict $G$ to capture $p_{data}$.

%\subsection*{A 1.4 DP property of $\mathcal{M}_p$}
%\textbf{Proposition 4.} Any G: possibly the best G an attacker can have in a communicational channel with quantized noise is DP.

%\textit{Proof} Let, any discriminator $ D = f(data)$ and $G = g(D,z)$, where $z$ is the noise. From the post-processing property in~\cite{DP-1} if $f(data)$ is DP we have $g(f(D))$ is DP. Also, given $D$ is DP, then $G$ is DP. Furthermore, if the training process is based on original data, FL-GAN follows DP~\cite{DP-2}. Now assume, the received $D$ is followed by quantized channel noise. From~\cite{AAAI-DP}, the received $D$ is DP. Hence, $G$ is DP as well.

%\textbf{Proposition 5. }$\mathcal{M}_p$ preserves privacy compared to full data sharing.

%\textit{Proof} Consider the mutual information between data, $D$ : $I(data,D)$, and data, data: $I(data,data)$. Therefore, we have $I(data,D) \leq I(data,data)$. Therefore, if we preserve some privacy in $D$ implies privacy in $data$.

\section*{C Results}

In this section, we provide more results and evaluations for PS-FedGAN.
\subsection*{C 1.1 Utility}
\subsubsection*{C 1.1.1 Classification Accuracy}
\begin{table}[h]
\centering
\caption{Classification accuracies of existing FL methods and PS-FedGAN in Split-3.  We use the following
abbreviations. FedAvg: FA, FedProx: FP, SCAFFOLD: SD, Naivemix: NM, FedMix: FM, SDA-FL:
SL, PS-FedGAN: PS, Sharing full cGAN: FG}
\begin{tabular}{lll l l}
\toprule
 % \multicolumn{1}{c}{Split }     & 1 & 2 & 3            \\
 %    \cmidrule(r){2-5}
Method & MNIST  & FMNIST & SVHN  & CIFAR10\\ 
\midrule
FA &  98.42 & 82.47 & 84.18 & 79.33\\ 
FP & 98.38 & 83.43 & 92.15&79.54\\ 
SD & 96.89 & 77.68 & 80.13 & 79.35 \\ 
NM & 98.11 & 82.09 &92.30 & 78.92\\ 
FM & 98.46 & 84.65& 92.61 & 79.49\\ 
SL & 98.50 & 87.06 & 93.16 & 84.56 \\ 
FG & 98.46 & 89.26&93.05& 82.09\\
\midrule
{PS} & {98.54} & {89.28} & {93.23} & {82.05} \\ 
 \bottomrule
\end{tabular}
\label{tab:2}
\end{table}

Table~\ref{tab:2} presents the classification accuracies of the existing FL methods compared with our proposed PS-FedGAN for Split-3. We observe a similar trend as we have seen from the previous results in Table~\ref{tab:1}. For the dataset MNIST, FMNIST, and SVHN, we keep the 1\% original data available on the server. For the CIFAR10, we consider 10 \% of the original data at the cloud server. Considering that other evaluated methods may have more original data to train the classifier, this ratio is reasonable in Split-3.
Shown as the results, our the proposed methods show superior and competitive performance compared to most of the SOTA methods which match the previous conclusions in Split-1/2.

\subsubsection*{C 1.1.2 Convergence of Cloud Generators}

\begin{figure}[b]
  \centering
  \includegraphics[width=0.7\columnwidth]{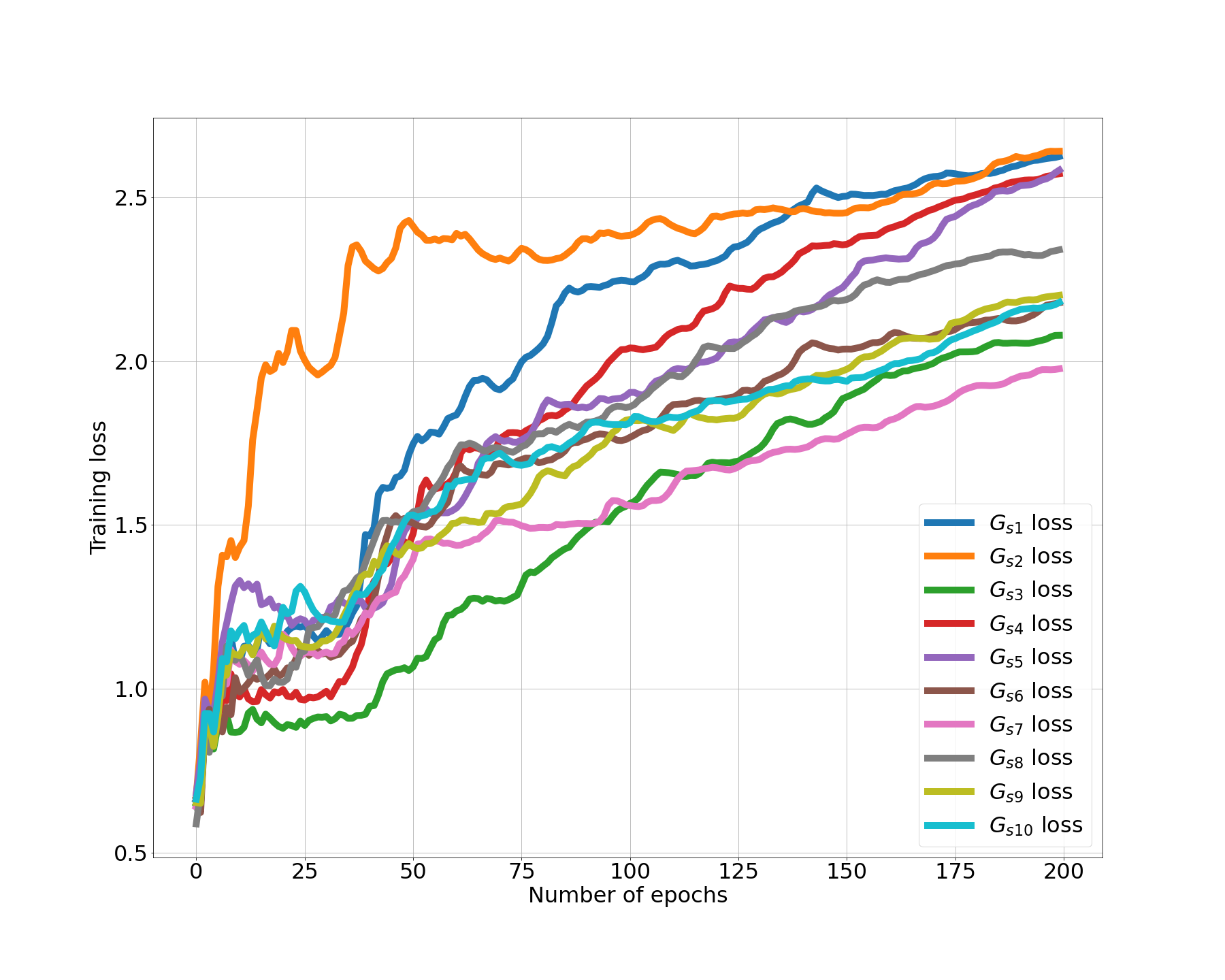}
  \caption{Generator loss of all the cloud generators in Split-1 for MNIST dataset. }
  \label{fig:cl1}
\end{figure}

As an example to illustrate the generator convergence, we plot the cloud generator loss of all the 10 users under Split-1 in Figure~\ref{fig:cl1}. Shown as the results, our proposed PS-FedGAN can converge well for all users, which further demonstrate the practicality of the discriminator sharing.

\subsubsection*{C 1.1.3 Divergence of Attacker's Generator}

\begin{figure}
  \centering
  \includegraphics[width=0.7\columnwidth]{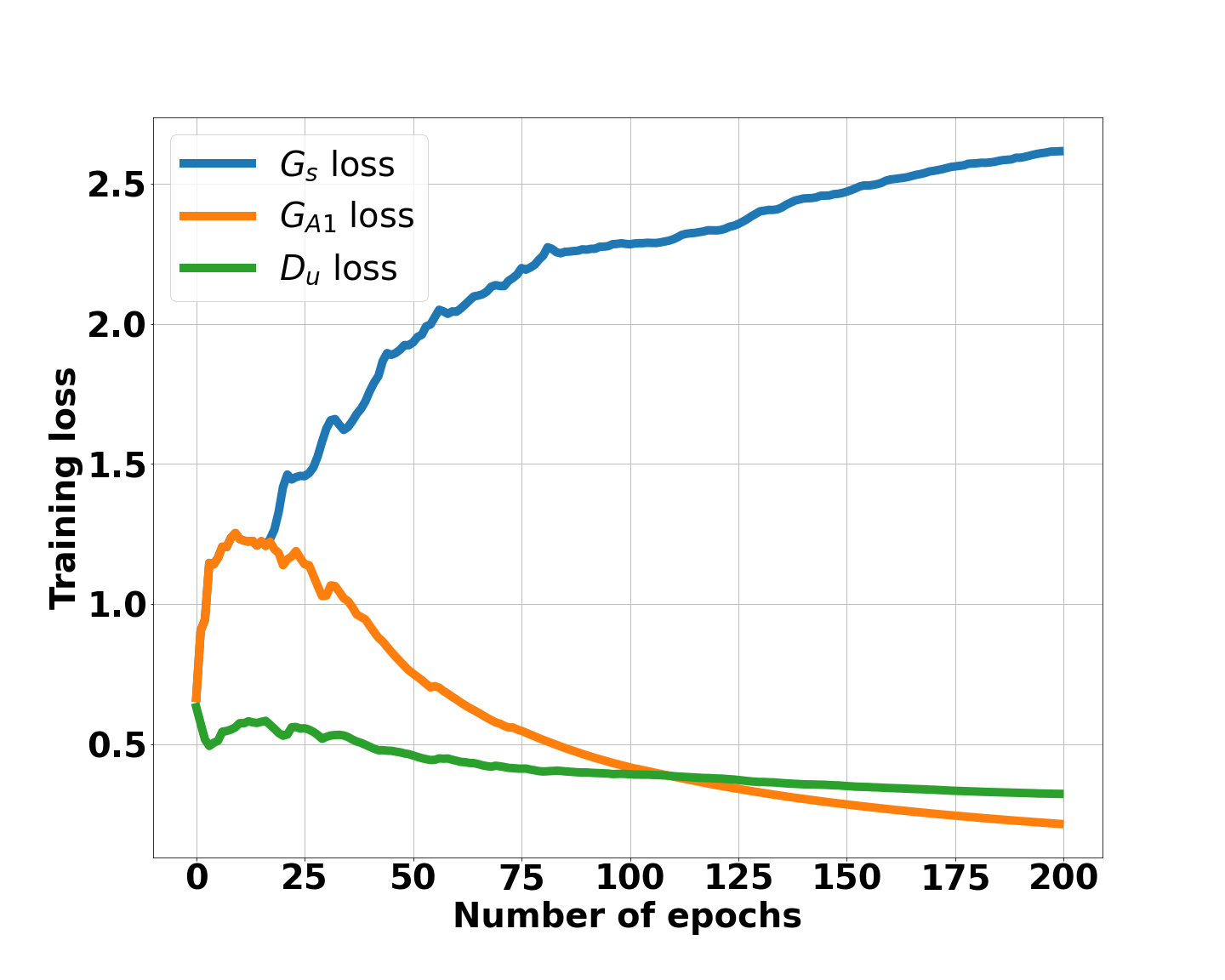}
  \caption{Divergence of the attacker. Any attacker fails to initiate with the same parameters as the optimal generators, i.e., $G_u$ or $G_s$, which indicate the attacks would fail to capture the local user's data distribution.  }
  \label{fig:div}
\end{figure}

Shown in Figure~\ref{fig:div} we see that if the attacker ($\mathcal{A}_1$) deviates from the actual generator weights, the convergence of the attacker would be to a trivial point. Hence, almost no underlying user data properties are captured by the attacker.

\subsection*{C 1.2 $\mathcal{A}_1$ Performances}
\begin{figure}
  \centering
  \includegraphics[width=\columnwidth]{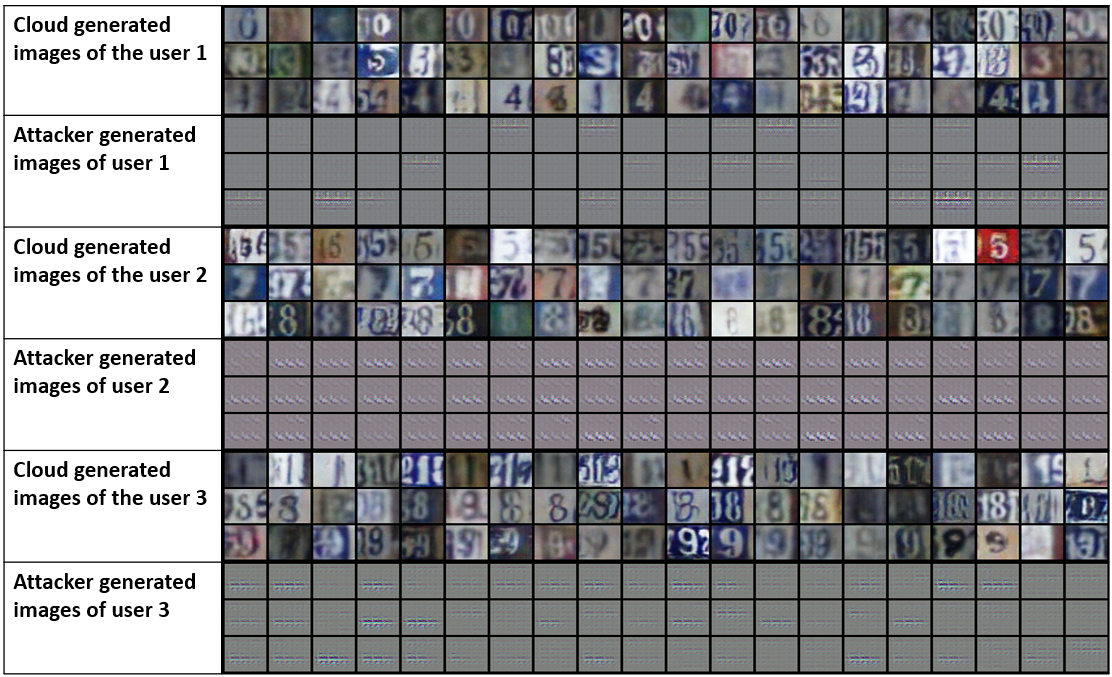}
  \caption{Cloud generated images compared to $\mathcal{A}_1$ generated images for Split-3 for SVHN dataset. }
  \label{fig:a1}
\end{figure}

\begin{figure}
  \centering
  \includegraphics[width=\columnwidth]{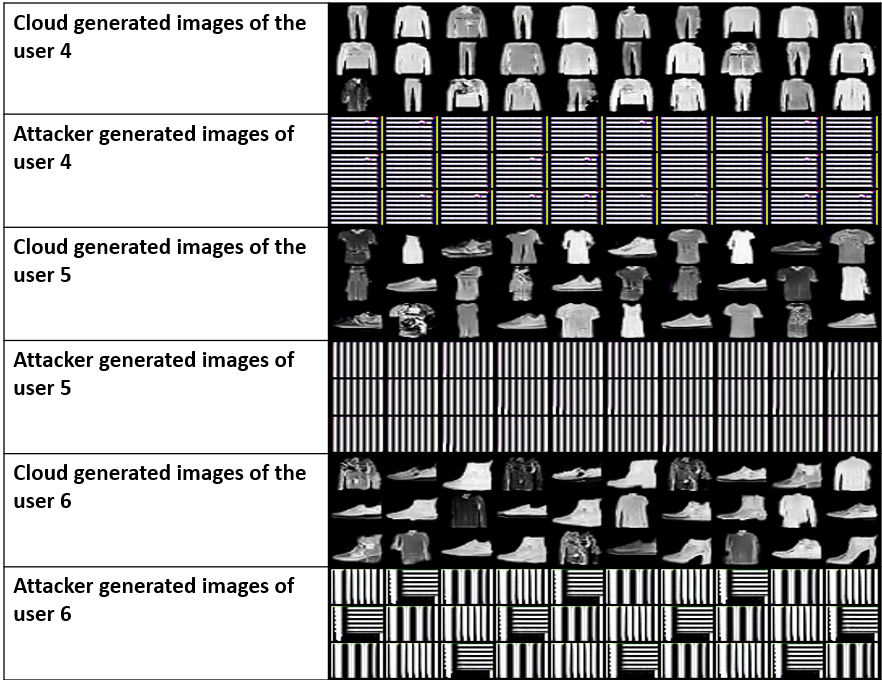}
  \caption{Cloud generated images compared to $\mathcal{A}_1$ generated images for Split-3 for FMNIST dataset. }
  \label{fig:a2}
\end{figure}

In this section, we provide more visual results for the robustness of the proposed methods against the attacks of $\mathcal{A}_1$. Figure~\ref{fig:a1} illustrates the cloud-generated images compared to $\mathcal{A}_1$ generated images for the SVHN dataset for three users at some intermediate step (epoch 20) while
Figure~\ref{fig:a2} show the regenerated samples in the FMNIST dataset. 
Shown as the visualized results, $\mathcal{A}_1$ fails to capture the underlying data statistics as well as to regenerate the meaningful synthetic data samples.

\end{document}